\documentclass[10pt,twocolumn,letterpaper]{article}

\usepackage{iccv}
\usepackage{times}
\usepackage{epsfig}
\usepackage{graphicx}
\usepackage{amsmath}
\usepackage{amssymb}

\graphicspath{{./figures/}}

\usepackage{xcolor}

\usepackage[pagebackref=true,breaklinks=true,letterpaper=true,colorlinks,bookmarks=false]{hyperref}

\iccvfinalcopy 

\def\iccvPaperID{1779} 

\ificcvfinal\fi
\begin{document}

\title{Neural Ctrl-F: Segmentation-free Query-by-String Word Spotting in Handwritten Manuscript Collections}

\author{Tomas Wilkinson\\
Department of Information Technology\\
Uppsala University\\
{\tt\small tomas.wilkinson@it.uu.se}
\and
Jonas Lindstr{\"o}m\\
Department of History\\
Uppsala University\\
\and
Anders Brun\\
Department of Information Technology\\
Uppsala University\\
}
\maketitle
\thispagestyle{empty}

\begin{abstract}
	In this paper, we approach the problem of segmentation-free query-by-string word spotting for handwritten documents. In other words, we use methods inspired from computer vision and machine learning to search for words in large collections of digitized manuscripts. In particular, we are interested in historical handwritten texts, which are often far more challenging than modern printed documents. This task is important, as it provides people with a way to quickly find what they are looking for in large collections that are tedious and difficult to read manually. To this end, we introduce an end-to-end trainable model based on deep neural networks that we call Ctrl-F-Net. Given a full manuscript page, the model simultaneously generates region proposals, and embeds these into a distributed word embedding space, where searches are performed. We evaluate the model on common benchmarks for handwritten word spotting, outperforming the previous state-of-the-art segmentation-free approaches by a large margin, and in some cases even segmentation-based approaches. One interesting real-life application of our approach is to help historians to find and count specific words in court records that are related to women's sustenance activities and division of labor. We provide promising preliminary experiments that validate our method on this task.
\end{abstract}

\section{Introduction}
\begin{figure}[t!]
	\begin{center}
		\includegraphics[width=0.97\linewidth]{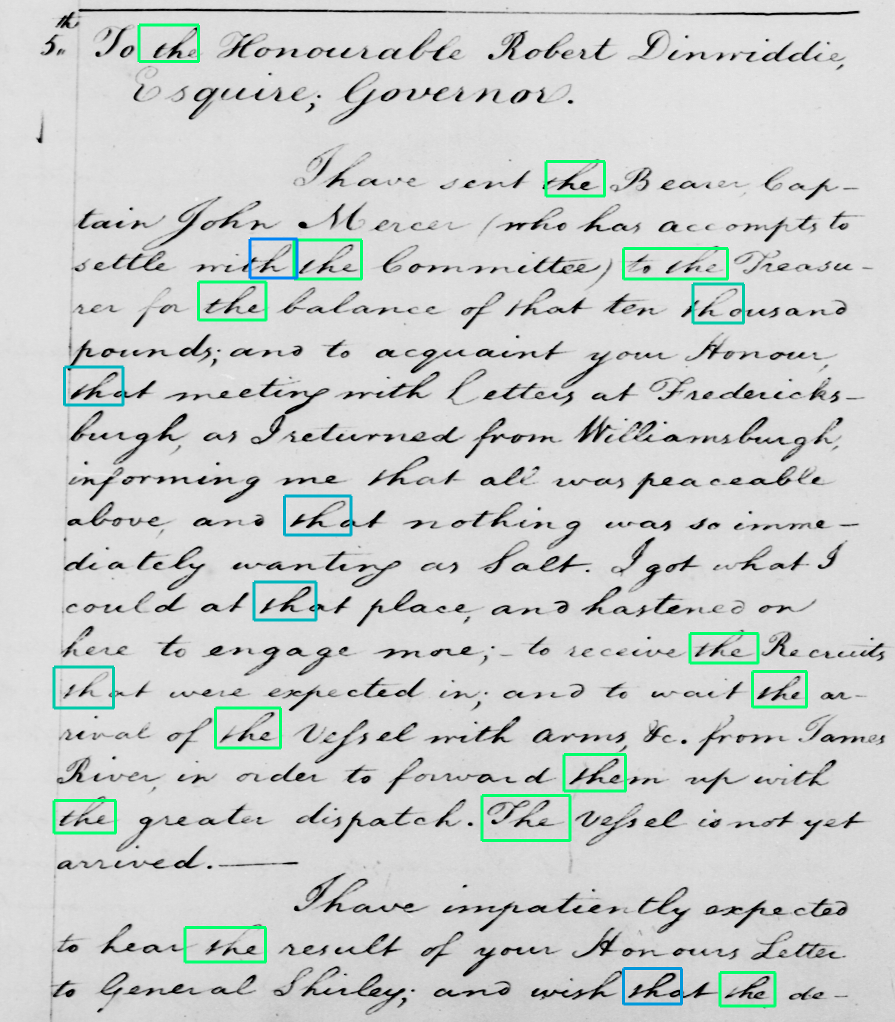}
	\end{center}
	\caption{A search for the word "the" on one page of the George Washington letters to Congress dataset, displaying the top 20 results. The greener the bounding box, the closer it is to the search query. Note that the true number of occurrences of the word "the" is 13 and the rest of the displayed results are parts of words that are similar to the query.}
	\label{fig:example_search}
\end{figure}

Word spotting is the task of searching through a collection of scanned manuscripts
to locate a provided search query, see Figure \ref{fig:example_search}. Word spotting gives an important oppertunity for digital humanities researchers, whose work is limited by the time spent on manually sifting through old manuscripts to find what they are looking for. Researchers working with old manuscripts can spend several months with a single book of a few hundred pages. There exist many digitized collections that are used for research purposes \cite{old_bailey, GenderAndWork, vaticanLibrary}, and although there have been work automatically mining these for information \cite{Hitchcock_Turkel_2016, Hitchcock_Turkel_2014, pettersson2015ranking, pettersson2016histsearch}, they are exclusively text-based methods. This means that the only way to search for information or do statistics in a collection is to painstakingly read and transcribe the manuscripts manually.

Although there have been successful crowdsourcing projects to transcribe manuscript collections \cite{causer2012building, oldweather}, they are limited to fairly modern transcripts and written in languages that are relatively common. Using the same approach for more esoteric work, written in multiple rarely spoken or dead languages and different alphabets and no canonical spelling would prove difficult. Similarly, OCR software have been used in research using documents, but OCR technology is even more limited than crowdsourcing, typically requiring neat machine printed text to produce legible results. In addition to these existing technologies, word spotting provides new possibilities for quantitative research in fields as diverse as demography, linguistics, paleography, genealogy, and history.

\begin{figure*}[t!]
	\begin{center}
		\includegraphics[width=0.99\linewidth]{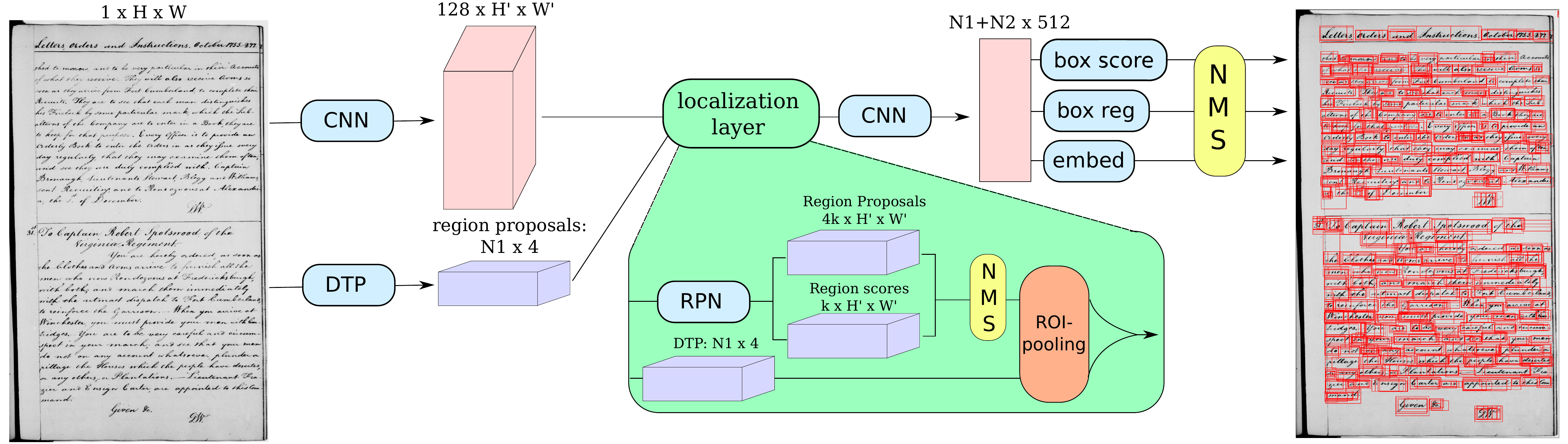}
	\end{center}
	\caption{The Ctrl-F-Net model. Given an input image, it is fed through the first CNN of the model and Dilated Text Proposals (DTP) are extracted. These are then fed into the localization layer, where additional text proposals are extracted using a Region Proposal Network (RPN), followed by non-max suppression. The RPN proposals are added to the DTP proposals and fed through the ROI-pooling layer, giving fixed length descriptors for each proposal. The proposals are fed through a second CNN and finally, each box coordinates are fine-tuned, given a wordness score, and a descriptor is extracted. Finally, a second non-max suppression is applied resulting a large number of region proposals, designed to replace the ground truth bounding boxes that are typically used.}
	\label{fig:ctrlf_net}
\end{figure*}

Compared to letter-by-letter text recognition, word spotting is a simpler task that, as a result, often is more transferable between sources (e.g., no language model is typically used); less data is required, which is crucial since manual annotation of historical manuscripts is very expensive as it often requires expert knowledge, making popular crowdsourcing alternatives (like amazon turk) not applicable; and since word spotting is designed to be more like a tool to find what you are looking for, manual inspection is typically done in any case. 

The task of word spotting is typically defined in ways that differ in two regards. The first is whether to do segmentation-based or segmentation-free word spotting. For segmentation-based word spotting, you assume that you have access to segmented word images, which is an unrealistic assumption when it comes to a real-life practical setting. This is not the case for segmentation-free word spotting, where you only need the image of a manuscript page. The second is whether or not the query is a manually cropped images of a word, query-by-examples (QbE), or a string of characters, query-by-string (QbS). Both work in a practical setting, though QbS is most often the preferred option since it does not require you to find an example of what you are looking for before you can search for more occurrences. Therefore, the preferred paradigm of word spotting is almost always segmentation-free QbS word spotting.

This work presents an end-to-end trainable model for segmentation-free query-by-string word spotting that is designed with the intent helping professionals and enthusiasts that work with manuscripts on a daily basis. Based on a deep Convolutional Neural Network (CNN) \cite{krizhevsky2012imagenet, he2016identity}, and using the recently introduced, fully differentiable dense localization layer for region-based training and prediction \cite{densecap}, together with the state-of-the-art word embedding approaches for handwritten word spotting \cite{wilkinson2016semantic}, we present a model for word spotting that outperforms existing state-of-the-art results on common benchmark datasets, as well as experimental results on an early 17\textsuperscript{th} century manuscript.

\section{Related Work}
Recently, segmentation-based word spotting has seen major improvements, primarily due to two advancements: distributed word representations, or word embeddings, and deep learning \cite{almazan2014word, sudholdtPhocnet, wilkinson2016semantic, KrishnanDeepFeatureEmbedding}. Although these advancements have been partially adapted to the segmentation-free setting \cite{ghosh2015query, Ghosh_word_spotting}, combining them has not yet been done. The most widely adopted embedding is the Pyramidal Histogram of Characters (PHOC) that has, since its introduction, become widely adopted for word spotting \cite{almazan2014word, sudholdtPhocnet, ghosh2015query, Ghosh_word_spotting, KrishnanDeepFeatureEmbedding} and work in lexicon-based text recognition \cite{poznanski2016cnn}.

Depending on how they generate regions from where to retrieve within the image, segmentation-free word spotting can often be grouped into two categories. The first category is sliding window based methods \cite{almazan2014segmentation, Ghosh_word_spotting, rothacker2015segmentation, rothacker2013bag}, where regions are generated at each position (either pixel or on a grid of dense SIFT or HOG features) in the manuscripts. These methods are typically used for QbE word spotting, where the size of the query in pixels is known, which allows for limiting the sizes of generated regions \cite{almazan2014segmentation, Ghosh_word_spotting,  rothacker2013bag}. To do QbS, you typically need to estimate the size of the region given the query string \cite{rothacker2015segmentation}. The main drawback with these methods is the large amounts of regions generated, resulting in false positives and long processing times \cite{kovalchuk2014simple}. Furthermore, due to the lack of attention, sliding window techniques are sensitive to small shifts in the input, which in turn shifts all the regions extracted in the image.

The second category consists of methods based on connected components \cite{ghosh2015query, kovalchuk2014simple}. These methods are typically based on binarizing the manuscript image, extracting connected components, and then grouping them in a bottom-up fashion following some heuristics, and then extracting bounding boxes. A similar approach is used in \cite{krishnan2016matching} for matching entire documents using distributions of word images. While still producing too many regions, they are fewer compared to methods based on the sliding window and they are not sensitive to shifts in the input to the same extent.

End-to-end scene text localization and recognition has recently received some of attention \cite{jaderberg2016reading, neumann2016real}, and the problem is similar to segmentation-free word spotting. In \cite{jaderberg2016reading}, an end-to-end system for text localization, recognition and retrieval based on region proposals and deep convolutional neural networks. A different approach is done in \cite{neumann2016real}, where individual characters are detected and compounded together in a bottom-up fashion to build words and text lines. Casting text recognition as text retrieval, \cite{rodriguez2015label} uses a label (word) embedding similar to PHOC to perform segmentation-based scene text recognition.

\section{Ctrl-F-Net}
The model (dubbed Ctrl-F-Net after the shortcut for word search in certain word processors) is inspired by previous work on object detection \cite{ren2015faster}, dense image captioning \cite{densecap} and segmentation-based word spotting \cite{wilkinson2016semantic}, and it is similar in spirit to the scene text recognition approach of \cite{jaderberg2016reading}. It allows for simultaneously proposing and scoring region proposals, and embedding the best proposals into the word embedding space, in which the search is performed. A total of five loss functions are used and the model is trained in an end-to-end manner allowing the model to attune to all the tasks at hand. An overview of the model can be seen in Figure \ref{fig:ctrlf_net}.

An input image is first fed through several layers of a CNN, until it has been downsampled in spatial size by a factor of 8. The 34-layer pre-activation ResNet \cite{he2016identity} was chosen as a CNN architecture due to its high performance and expressiveness compared to its small memory footprint. The feature maps are then fed through a Region Proposal Network (RPN) \cite{ren2015faster,densecap}, which attempts to regress $K=15$ anchor boxes of different aspect ratios and sizes at each spatial position on the feature maps in a sliding window fashion. The boxes are also given a \emph{wordness} score, whether or not they are situated atop a word, and a non-max suppression step is applied to said score. Proposals with an Intersection-over-union (IoU) overlap greater than 0.75 are considered as positives and 128 are randomly sampled, along with 128 random negatives (boxes with IoU less than 0.4) calculate the wordness and regression losses.

The 256 sampled boxes are then fed through a ROI-pooling layer \cite{girshick2015fast, densecap} with Bilinear Interpolation \cite{jaderberg2015spatial, densecap} that gives a fixed-length output for the input boxes that vary in size. These are then fed through the rest of the CNN and then used as input to three branches. The first regresses the box coordinates once again, refining them further. The second gives a final wordness score, and the third is a small fully-connected embedding network that embeds the words into the word embedding space.

In the output layer and the localization layer, we use a binary logistic loss for confidence scores of whether or not a region is positive. The bounding boxes are parameterized according to \cite{girshick2015fast}, both for the anchor box regression and the output box regression. The boxes are represented as the quadruples $(x_c, y_c, w, h)$, where $x_c$ and $y_c$ are the center of a box and $w$ and $h$ is its width and height. The functions to learn are normalized translation offsets for $x$ and $y$ and log-space scaling factors for $w$ and $h$. The loss function is a smooth $l1$ loss
\begin{equation}
L_{reg}(x_i, t_i) =
\left\{
  \begin{array}{lcr}
  	0.5 \cdot (z_i)^2  &\textnormal{if }  |z_i| & < 1 \\
  	|z_i| - 0.5  &\textnormal{if }  |z_i| & \geq 1 \\
  \end{array}
\right.
\end{equation}
where $z_i = x_i - t_i$, $x_i$ is one of $(x_c, y_c, w, h)$ and $t_i$ is its corresponding target.

The embedding branch is a fully-connected network with two hidden layers of size 4096, with batch normalization\cite{ioffe2015batch} after each layer and the hyperbolic tangent is used as activation function. The final layer is an $l^2$-normalization layer. It only receives the regions labelled positive as input. Following \cite{wilkinson2016semantic}, the cosine embedding loss function is used to learn the embeddings, and is defined as
\begin{equation}
L_{emb}(\mathbf{u}, \mathbf{v}) =
\left\{
  \begin{array}{lcr}
     1 - \mathbf{u}^\mathsf{T} \mathbf{v} &\textnormal{if }  y = 0 \\
    \max(0, \mathbf{u}^\mathsf{T} \mathbf{v} - \gamma) &\textnormal{if }  y = 1
  \end{array}
\right.
\end{equation}
where $\mathbf{v}$ is an embedding of a positive region and $\mathbf{u}$ is the embedding of the ground truth label with which the region overlaps the most. The label $y$ is set to 1 if the label embedding and the region matches, in which case they are moved closer together, and set to 0 if they do not match wherein they are moved apart. The margin $\gamma$ is fixed at 0.1 for all the experiments.

The final loss function is a weighted linear combination of the five losses. The weights used for all the experiments are 0.01 for the RPN losses, 0.1 for the output box regression and scoring and 3 for the cosine embedding loss.

\subsection{Querying}
During testing, DTP proposals and the manuscript pages are fed through the CTRL-F-net, which outputs $N1 + N2$ proposals region proposals, their wordness scores, and their corresponding descriptors. We then do a thresholding on the wordness score, where we remove region proposals that are below a threshold, which we keep fixed a 0.01 for all experiments. We set $N2 = N1$ on a page-by-page basis to evaluate the two proposal generators fairly. When a query is selected, either by cropping a part of an image and feeding it through the model for QbE retrieval or by providing a search query for QbS, it is first transformed to the word embedding space. The query is then compared by means of the cosine distance to each region proposal and sorted the w.r.t. their similarity to the query. Then a third and final non-max suppression step is applied using the region proposals similarity to the query word as a score.

\begin{figure}[t!]
	\begin{center}
		\includegraphics[height=0.37\linewidth]{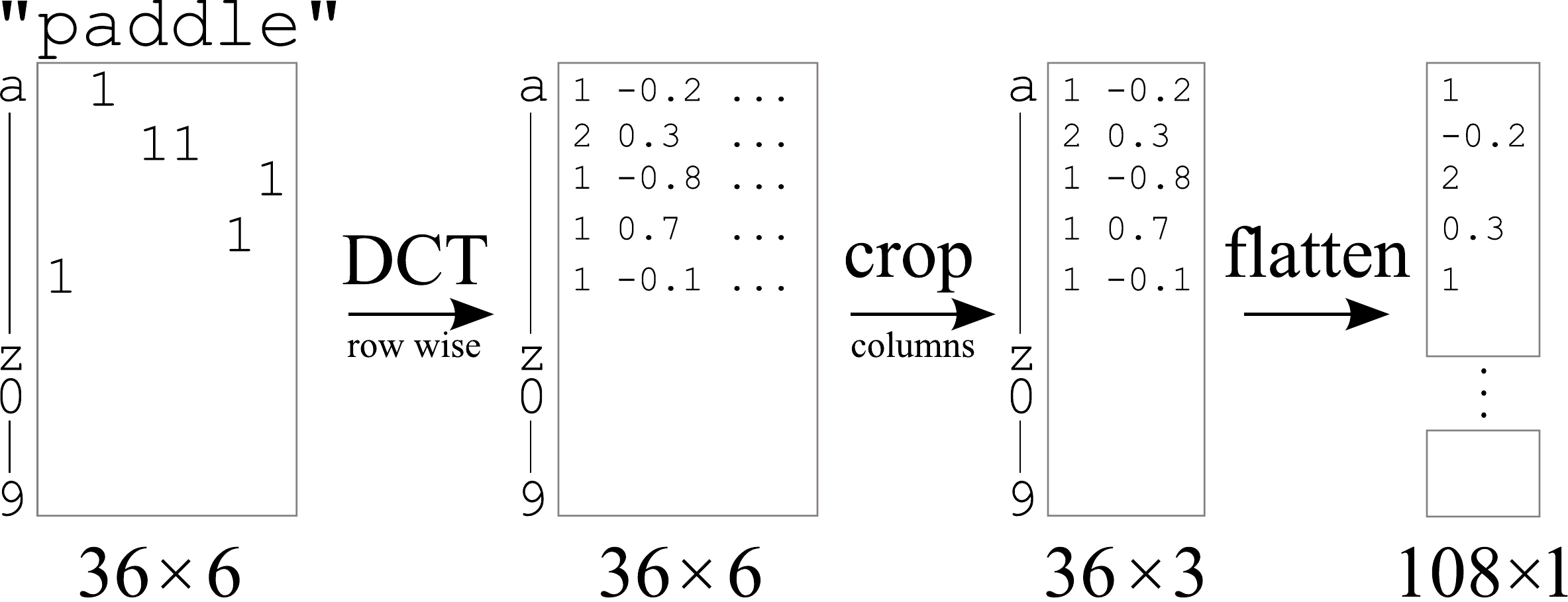} \\
		\vspace{0.5cm}
		\includegraphics[height=0.37\linewidth]{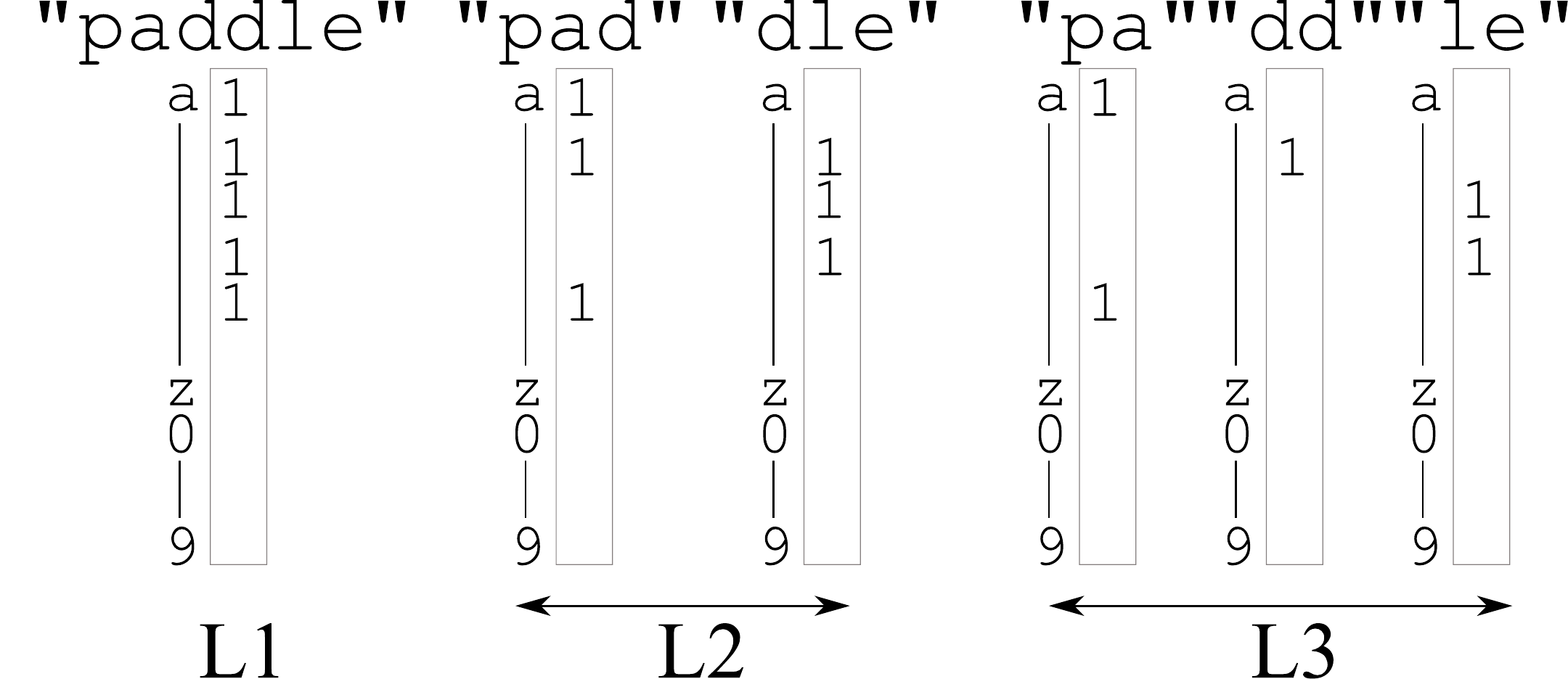}
	\end{center}
	\caption{The two word embeddings evaluated in this paper, DCToW (top) and PHOC (bottom). Note that we only show 3 of the 5 levels of the PHOC embedding here. The final PHOC dimensionality is $36 \cdot (1 + 2 + 3 + 4 + 5) = 540$.}
	\label{fig:embeddings}
\end{figure}

\subsection{Word Embeddings}
We evaluate two different word embeddings that have previously shown good performance in the segmentation-based setting, see Figure \ref{fig:embeddings}. The first embedding is the Discrete Cosine Transform of Words (DCToW), recently introduced in \cite{wilkinson2016semantic}. It is a low frequency, distributed representation of a word, that has recently achieved state-of-the-art results in segmentation-based word spotting. Given a word of length $m$ and an alphabet of length $K$ (we use the digits 0-9 and lower-case letters for all experiments), each character is first transformed to a one-hot representation vector. These vectors are concatenated into a $m \times K$ matrix for the whole word. Then, a Discrete Cosine Transform is applied to each row of the matrix. Finally, this matrix is cropped, keeping only $r$ first low-frequency components and those are flattened into a vector. For all experiments in this paper, $r$ is set to $3$, which results in a $3 \cdot K = 3 \cdot 36 = 108$ dimensional vector. Words that are shorter than $r$ characters are padded with zeros to get the correct length.

The second is a version of the well established Pyramidal Histogram of Characters (PHOC), that has achieved good success in the last few years \cite{almazan2014word, KrishnanDeepFeatureEmbedding, sudholdtPhocnet, poznanski2016cnn}. Given a word, an alphabet, and a number of pyramid levels, the first step is calculating a binary occurrence vector, which encodes whether or not a character is present in the given word. Then the word is split into two and another binary occurrence vector is calculated for each of the two sub-words. This continues for the number of pyramid levels, in this case $5$, all of which are then concatenated length wise, resulting in a $36 \cdot (1 + 2 + 3 + 4 + 5) = 540$ dimensional binary vector. Previous work using the PHOC have made use of an additional $50$ of the most common English bi-grams to augment the alphabet, but we have found that it does not increase performance while at the same time increasing the complexity, hence they are omitted from the alphabet.

\subsection{Dilated Text Proposals}
High recall is an important part of a segmentation-free word spotting system. As the RPN is a sliding window approach, complementing its proposals with an external region proposal method based on connected components is likely to improve the recall rate. To extract external proposals, we use the recently introduced method from \cite{wilkinson2015novel}, which is specially designed for manuscript images, and can be considered to be based on connected components. While the method is not given a name, we call it Dilated Text Proposals (DTP) to increase the clarity of this work. Given a grayscale image, DTP first creates a set of $j$ binary image by thresholding at $j$ different multiples of the image mean value. Then apply morphological closing to each binary image using a set of $l$ generated rectangular kernels. For each of the $j \cdot l$ images, find the connected components, then extract bounding boxes for each connected component and remove duplicate boxes.

\section{Data Augmentation}
\begin{figure}[t!]
	\begin{center}
		\includegraphics[width=0.45\linewidth]{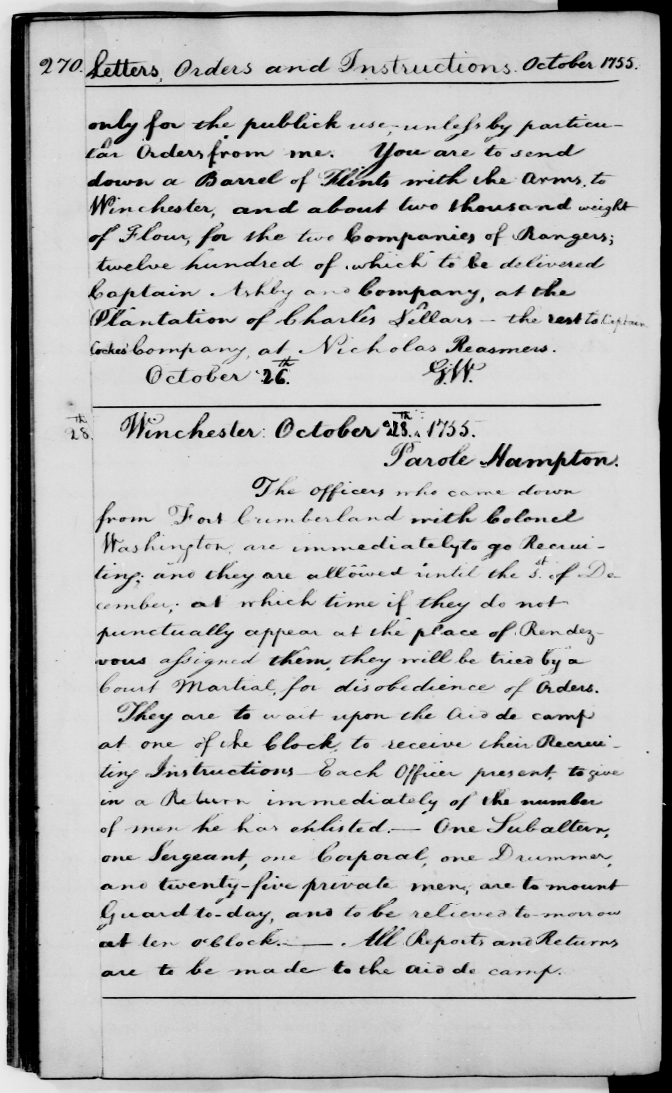}
	    \hspace{0.2cm}
		\includegraphics[width=0.45\linewidth]{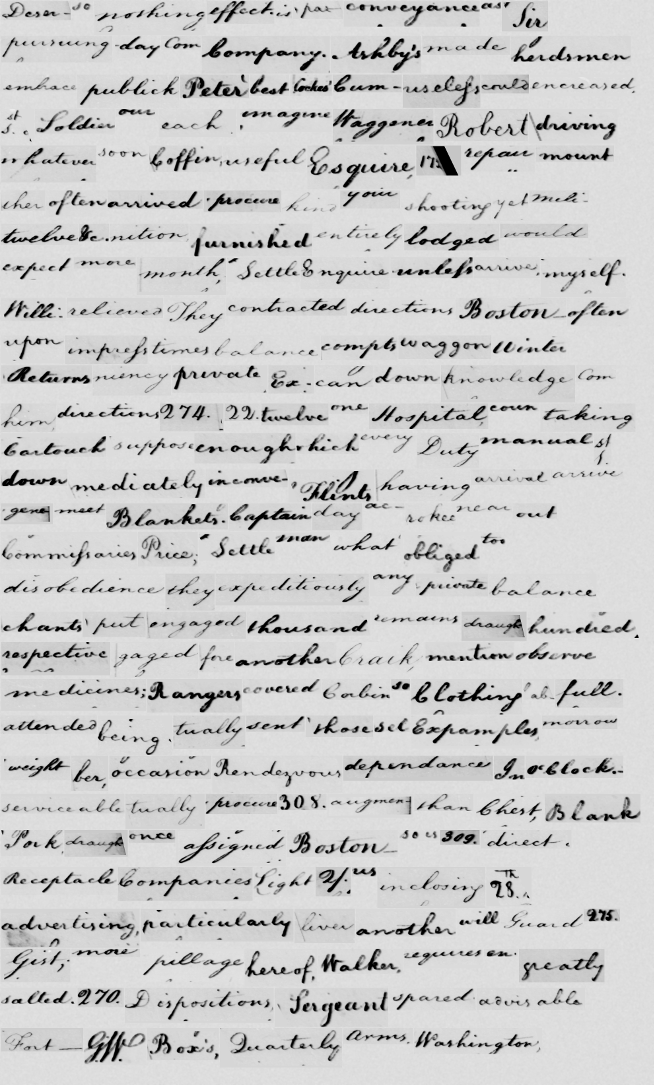}
	\end{center}
	\caption{Top: a visual comparison between in-place (left) and full page augmentation (right). In-place augmentation primarily regularizes the box regression and scoring while full page augmentation helps with learning discriminative word embeddings.}
	\label{fig:data_aug}
\end{figure}

To properly train the model, we make extensive use of two, slightly differing, data augmentation strategies, \emph{full-page} and \emph{in-place} augmentation. They provide complementary properties help the model to learn a discriminative word embedding and to generate high quality region proposal. Figure \ref{fig:data_aug} shows a comparison of the two methods.

The full-page augmentation is based on using the augmentation strategy for single images of words from \cite{wilkinson2016semantic} to augment word images and placing them row-by-row on a background canvas. The canvas is created by adding some Gaussian noise to a background color that is randomly sampled around the median of the image. The finished augmented page looks like left-aligned manuscripts of randomly sampled word images. This lets us completely control of the distribution of classes, and where to place them, making it possible to get a uniform class distribution, which greatly helps the learning. The augmentation is an affine transform (most importantly a shearing transformation) followed by a grayscale morphological dilation or erosion, either fattening or thinning the ink.

The in-place augmentation uses the same basic word-level augmentation but is designed to keep the overall look of the manuscripts intact. Given an image of a manuscript page, we iterate through each of the ground truth bounding boxes and augment each word in-place, with a shearing transform followed by a gray-scale morphological dilation or erosion, while ensuring the same size of the word image so that it can be slotted back into the page. This augmentation helps the model generate and score the region proposals.

\section{Experiments}
We evaluate our model on three datasets. Two common word spotting benchmarks and one historical manuscript from the early 17\textsuperscript{th} century. The first is the George Washington (GW) dataset \cite{lavrenko2004holistic}  written in English the middle of the 18\textsuperscript{th} century by George Washington and his secretaries, it consists of 20 pages, or 4860 words. We follow the evaluation procedure used in \cite{rothacker2015segmentation}, by splitting the pages into a training and validation set of 15 pages, setting aside 5 for testing, and also doing a 5-15 split of train/val and testing. In both cases, we use 1 page as a validation set. The results reported is the average of four cross validations.

The second dataset is the IAM offline handwriting dataset \cite{marti2002iam}, a modern cursive dataset consisting of 1539 pages, or 115320 words, written by 657 writers. We use the official train/val/test split for \emph{writer independent text line recognition}, where there is no writer overlap between the different splits. Following standard protocol, we remove stop words from the set of queries, and in line with \cite{almazan2014word}, queries that come from lines that are marked as containing segmentation errors are removed. Ground truth boxes that are so small that they collapse to a width or height of zero when downsampled by a factor 8 are also removed.

\begin{figure}[t!]
	\begin{center}
		\includegraphics[width=0.99\linewidth]{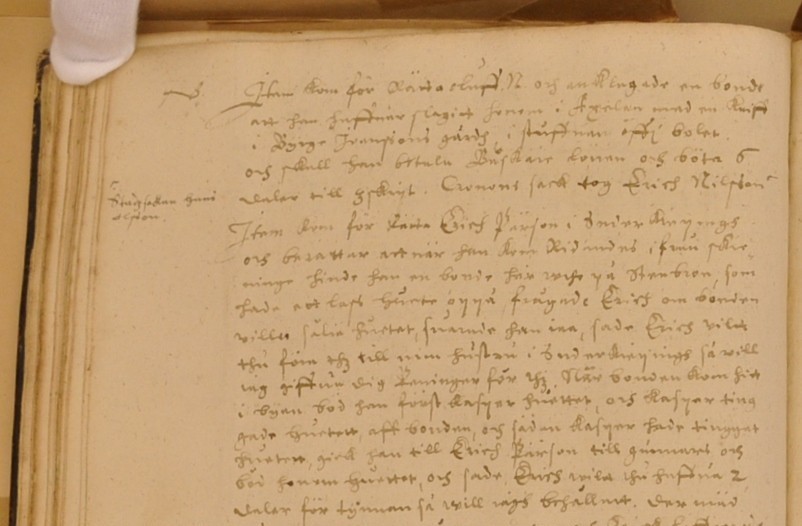}
	\end{center}
	\caption{An example image of the court records dataset.}
	\label{fig:dombok}
\end{figure}

The third dataset is the records of the magistrate court of the Swedish town Link{\"o}ping, written between 1609 and 1616. Because of their richness and the variety of information they contain, court records are much used in historical research. They are time consuming to work with, however, for the same reasons. The present volume is written in neo-gothic cursive script. It consists of 150 pages, or 34326 words. A part of a page can be seen in Figure \ref{fig:dombok}. We have access to page-level transcriptions of each page. We manually annotate 5 pages for training and 1 page for validation, and the rest is used for testing.

\begin{table*}[t!]
\renewcommand{\arraystretch}{1.3}
\caption{MAP comparison in \% with state-of-the-art segmentation-free methods on the GW dataset. The Ctrl-F-Net results marked with an asterisk use the evaluation protocol from \cite{Ghosh_word_spotting,ghosh2015query}.}

\label{tab:sota_comp_gw}
\centering
\begin{tabular}{|l|c|c|c|c|c|c|c|c|c|c|c|c|}
\hline
& \multicolumn{4}{c|}{GW 15-5}  & \multicolumn{4}{c|}{GW 5-15} & \multicolumn{4}{c|}{IAM} \\
\hline
& \multicolumn{2}{c|}{50\%} & \multicolumn{2}{c|}{25\%} & \multicolumn{2}{c|}{50\%} & \multicolumn{2}{c|}{25\%} & \multicolumn{2}{c|}{50\%} & \multicolumn{2}{c|}{25\%} \\
\hline
 & QbE & QbS & QbE & QbS & QbE & QbS & QbE & QbS& QbE & QbS & QbE & QbS \\
\hline
Ctrl-F-Net DCToW & 90.5 & \textbf{91.0} & \textbf{97.0} & \textbf{95.2} & \textbf{85.2} & \textbf{73.8} & \textbf{91.6} & \textbf{76.8}  & \textbf{72.0} & \textbf{80.3} & \textbf{74.1} & \textbf{82.5}\\
\hline
Ctrl-F-Net PHOC & \textbf{90.9} & 90.1 & 96.7 & 93.9 & 83.1 & 65.6 & 89.4 & 68.2 & 71.5 & 78.8 & 73.7 & 80.8 \\
\hline
BoF HMMs \cite{rothacker2015segmentation} & - & 76.5 & - & 80.1 & - & 54.6 & - & 58.1 & - & - & - & -\\
\hline
\hline
Ctrl-F-Net DCToW* & \textbf{79.7} & \textbf{90.4} & \textbf{95.1} & \textbf{96.3} &  - & - & - & - & - & - & - & -\\
\hline
SW PHOC \cite{Ghosh_word_spotting} & 67.7 & - & - & - & - & - & - & - & 42.1 & - & - & -\\
\hline
BG index \cite{ghosh2015query} & - & 73.3 & - & - & - & - & - & - & - & 48.6 & - & -\\
\hline
\end{tabular}
\end{table*}

\subsection{Training}
The model is trained end-to-end in a single phase. We first train a model (with weights initialized randomly) using the synthetic IIIT-HWS-10k dataset \cite{krishnan2016matching}. Since it only consists of word images, we use the full-page augmentation technique to create 5000 synthetic document images. This model is used to initialize all other models, except for the court records, where the model is initialized with a model trained on the GW dataset. For the other datasets, we create 5000 augmented images, split evenly between in-place and full-page augmentation, and add them to the original data. The input image is rescaled such that its longest side is 1720. We train each model for a maximum of 25000 iterations, and the measure the performance on a held out validation set every 1000 iterations. The model with the highest validation MAP score is used for testing. The learning rate is initially set to $2^{-3}$ and is multiplied every 10000 iterations by 0.1, we use ADAM \cite{kingma2014adam} to update the weights. 

\subsection{Evaluation}
For the GW and IAM datasets, we evaluate our model using the standard metric used for word spotting, Mean Average Precision (MAP), where the Average Precision is defined  as
\begin{equation}
AP = \frac{\sum_{k}^{N} P(k) \times r(k) }{|r|}
\end{equation}
where $P(k)$ is the precision measured at cut-off $k$ in the returned list and $r(k)$ is an indicator function that is 1 if a returned result at rank $k$ is relevant, and 0 otherwise. A retrieved word is considered relevant if its IoU overlap with a ground truth box is greater than a threshold $t_o \in \{0.25, 0.5\}$ and the label matches the query. The MAP score is the mean of the AP over the queries.
\begin{equation}
MAP = \frac{\sum_{q}^{Q} AP(q) }{|Q|}
\end{equation}
For the QbE evaluation, all the ground truth segmented word images in the test set is used. For QbS, all unique ground truth labels are used. We use a score nms overlap threshold of 0.4 and a query nms overlap threshold of 0.01.

For the court records, ground truth annotation on the word level does not exists. However, we have access to unaligned transcriptions of each image. Using this information, we redefine the relevancy, $r(k)$, of the average precision to whether or not a word image that is retrieved appears in the transcription of a page. To avoid counting overly many duplicates, the AP for a query is only calculated until we have retrieved regions from pages the same number of times they occur in the transcription. We report QbS results using a set of specially selected queries that are of significant interest to historians working with this and similar manuscripts. The queries concern female designators. Recent historical research has suggested that the vernacular word for "wife" had a wider meaning than just “married woman” in the early modern period  \cite{christopher2014vad}.  In order to test this hypothesis, more evidence is needed of how this word, and similar words were used.

\subsection{Results}
We evaluate the some different model choices and compare the performance of Ctrl-F-Net to the state-of-the-art in segmentation-free and segmentation-based handwritten word spotting for the IAM and GW datasets. We also report recall results of the RPN, DTP and their combination. For the court records, we report the modified QbS AP scores for the set of selected queries and some qualitative word retrievals.

Table \ref{tab:sota_comp_gw} shows the results on the GW dataset compared to the state-of-the-art. We compare the DCToW and PHOC embeddings, where the former outperforms the latter across the board by a small margin. When comparing to other methods, our model outperforms them by a large margin, both in QbE and QbS and with respect to the different data splits. The Ctrl-F-Net result marked with an asterisk uses the evaluation protocol of \cite{Ghosh_word_spotting, ghosh2015query}, where all word instances in the dataset are used as queries for QbE, and all unique labels for QbS. The search is performed in all 20 pages.

Similar trends are observed on the IAM dataset, where we outperform the previous state-of-the-art by a large margin. We note that the comparison to \cite{Ghosh_word_spotting} and \cite{ghosh2015query} on the IAM dataset is not directly comparable, as they do line spotting where whole lines are retrieved and they perform their search in the annotated text lines, not the full pages, and their distance between a query and a text line is the shortest distance between the query and the word candidates of that line. According to the results presented in \cite{almazan2014word}, this is a slightly easier task.

In Table \ref{tab:sota_seg_comp}, we compare the best segmentation-free setup with a 25\% overlap threshold with state of the art methods for segmentation-based word spotting, that use the same evaluation protocol. We observe that our segmentation-free results are very competitive compared to the best segmentation-based approaches, even though they depend on manually segmented bounding boxes. In the case of QbS on the GW dataset, we even outperform the best segmentation-based approaches.

Table \ref{tab:recall_comp} shows the recall rates of the two sources of region proposals, the RPN and the DTP, and their union using the DCToW embedding on 15-5 and 5-15 GW datasets and the IAM dataset. The recall is the average over pages, and the number of proposals for each method is the same.

\begin{figure}
	\begin{center}
		\includegraphics[height=0.99\linewidth]{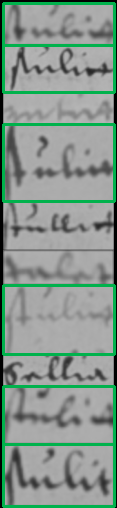}
		\hspace{0.1cm}
		\includegraphics[height=0.99\linewidth]{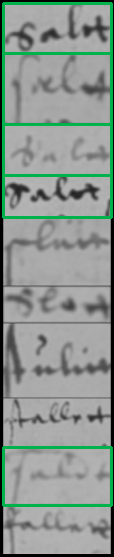}
		\hspace{0.1cm}
		\includegraphics[height=0.99\linewidth]{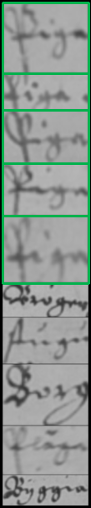}
		\hspace{0.1cm}
		\includegraphics[height=0.99\linewidth]{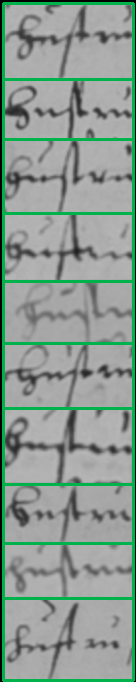}
	\end{center}
	\caption{Qualitative search results. The figure depicts the top 10 results starting from the top for the four queries "stulitt" (stolen), "saltt" (sold), "piga" (maid) and "hustru" (wife). Correct retrievals are highlighted in green. }
	\label{fig:qualitative_results}
\end{figure}

Table \ref{tab:dombok} shows the Average Precision of a few queries and the MAP av all the queries. We also show whether or not the queries are included in the training set. We include some qualitative results for the court record dataset in Figure \ref{fig:qualitative_results}, where the top 10 retrieved results for four queries is shown. 

\begin{table}
	\renewcommand{\arraystretch}{1.3}
	\caption{MAP comparison in \% with state-of-the-art segmentation-based methods using a 25\% overlap threshold, using the 15-5 page train/test split. Note that methods marked with $^\dagger$ use on pre-segmented word images.}
	\label{tab:sota_seg_comp}
	\centering
	\begin{tabular}{|l|c|c|c|c|}
		\hline
		Methods & \multicolumn{2}{c|}{GW} & \multicolumn{2}{c|}{IAM} \\
		\hline
		& QbE & QbS & QbE & QbS \\
		\hline
		Ctrl-F-Net DCToW & 97.0 & \textbf{95.2} & 74.1 & 82.5 \\
		\hline
		Embed attributes$^\dagger$ \cite{almazan2014word} &  93.0 & 91.3 & 55.7 & 73.7 \\
		\hline
		PHOCNet$^\dagger$ \cite{sudholdtPhocnet} &  96.7 & 92.6 & 72.5 & 84.0 \\
		\hline
		DCToW$^\dagger$ \cite{wilkinson2016semantic} &  \textbf{98.0} & 93.7 & 77.0 & 85.3 \\
		\hline
		Deep Embedding$^\dagger$ \cite{KrishnanDeepFeatureEmbedding} &  94.4 & 92.8 & \textbf{84.2} & \textbf{91.6} \\
		\hline
	\end{tabular}
\end{table}

\begin{table}
	\renewcommand{\arraystretch}{1.3}
	\caption{Recall comparison in \%, averaged over pages between the region proposal network and dilated text proposals using DCToW.}
	\label{tab:recall_comp}
	\centering
	\begin{tabular}{llccc}
		\hline
		Dataset & Overlap & RPN & DTP & Combined \\
		\hline
		GW 15-5 & 50\%& 91.1 & 98.1 & 99.4 \\
		\hline
		GW 15-5 & 25\%& 95.7 & 99.5 & 99.9 \\
		\hline
		GW 5-15 & 50\%& 82.3 & 96.7 & 97.5 \\
		\hline
		GW 5-15 & 25\%& 87.0 & 98.6 & 99.1 \\
		\hline
		IAM     & 50\%& 39.0 & 97.9 & 98.1 \\
		\hline
		IAM     & 25\%& 58.1 & 98.8 & 98.9 \\
		\hline
	\end{tabular}
\end{table}

\begin{table*}
\renewcommand{\arraystretch}{1.3}
\caption{Modified Average precision scores for a subset of the queries for the court record dataset.}
\label{tab:dombok}
\centering
\begin{tabular}{|l|c|c|c|c|c|c|c|c|c|c|}
	\hline
	Queries & hustru & quinna & piga & kona & stulitt & tient & tientte & hielpa & saltt &  mAP \\
	\hline
	English translation & wife & woman & maid & woman & stolen & earn & earned & help & sold &   \\
	\hline
	Ctrl-F-Net DCToW & 97.9 & 55.9 & 78.4 & 48.8 & 56.6 & 29.1 & 7.2 & 41.5 & 58.9 &  43.1 \\
	\hline
	True Occurrences & 114 & 8 & 7 & 3 & 10 & 3 & 2 & 3 & 7 & \\
	\hline
\end{tabular}
\end{table*}

\section{Discussion}
We can make several observations from the experiments. The first is that using a 25\% overlap threshold increases the result across the board, which is a bit surprising. This could be explained by the fact that the GW dataset is not very tightly labeled. This means that for single letter words like "I", the amount of space surrounding the ground truth box makes a region proposal that tightly attends to the ink overlapping less than 50\%. In a similar manner for the IAM dataset, the effect of the erroneous word level segmentations is less noticeable with the 25\% overlap threshold. For the purposes of the word search application, 25\% overlap would suffice as the user would in any case manually inspect each result.

A second observation, and a surprising one at that, is that the performance of the segmentation-free word spotting is competitive with the state-of-the-art segmentation-based methods, and in the case of QbS on GW, it outperforms all the segmentation-based methods. A possible explanation for this is that the region proposal methods are more consistent in how they are generated compared to the manually segmented ground truth boxes, which contain a random amount of space around each word. 

Furthermore, the results show that the Ctrl-F-Net is robust with respect to parameterization, since the same hyper parameters are used for all experiments. It is also robust with respect to multiple writers, as shown on the IAM dataset. The results using limited data, 4 training pages for GW and 5 for the court records, are impressive and possible due to the extensive use of data augmentation. Although achieving a good recall rate by itself, there is still a substantial gain in recall to be made by adding the external DTP to the RPN proposals. We also note that we have introduced two new kinds of data augmentation for documents: in place augmentation and full page augmentation. These can certainly be developed further, but the current implementation served its purpose for the this paper. 

The DCToW embedding achieves higher MAP scores compared to the PHOC in 11 out of 12 experiments, which could be taken as evidence that the DCToW is more robust in our segmentation-free setting. However, we cannot conclude that DCToW is better than PHOC in general. The performance in our setting could be an effect of our use of a cosine embedding loss function. In \cite{sudholdtPhocnet} for example, they estimate PHOC embeddings by using a binary classifier for each dimension of the PHOC.

We provide preliminary results on a novel real-life application, where we search for specific queries of interest in challenging 17\textsuperscript{th} century court records. Research in historical documents often consists of manually searching for small and scattered pieces of information in large amounts of texts. Only to find where to look in a book, or even which book to examine, can be time consuming; it is a matter of looking for a needle in a haystack. In effect, a historians interpretations are based on limited sets of data while other inquiries cannot be conducted at all. Speeding up the process of identifying relevant sections in handwritten texts would not only make it possible to gather more data, but would make way for new questions to be researched, as well. In particular, quantitative and statistical investigations for a large document collection.

\subsection{How many RPN Proposals?}
A crucial parameter to get the RPN to achieve a high recall is how many regions to propose. For tasks like object detection and dense image captioning on Imagenet \cite{russakovsky2015imagenet} and MSCOCO \cite{lin2014microsoft}, the typical number of proposals is 300 \cite{ren2015faster, densecap}. However, the full page manuscript images differ in two important regards, image size and the number of instances to detect. In order to ensure that the text is legible, manuscript images are typically captured in high resolution and cannot be downsampled by much. For this paper, images are resampled such that their largest side is $1720$ pixels. Overall, the images are much larger than their counterparts in datasets like Imagenet and MSCOCO, and thus presumably, requiring more region proposals.

MSCOCO has on average 7.7 object instances per image. For Imagenet and PascalVOC, this number is even lower, less than 3. Compare this to the GW dataset, which has 230 words per image, approximately 30 times greater than MSCOCO. Scaling up 300 proposals to 230 words per image gives us $300 \cdot 30 = 9000$ proposals. Adding this estimate to the image size difference, it is uncertain as to how many proposals you should expect to need. Using an equal number of RPN and DTP proposals, around 13200 per page on the GW dataset, the DTP has higher recall than the RPN (Table \ref{tab:recall_comp}), justifying its use. This is even more evident on the IAM dataset, which has approximately 1400 proposals per page on average. As the RPN is a sliding window based approach, it suffers from the common issue that more regions are needed to achieve high recall. A conclusion that could be drawn from this is that more diverse image data (medical, manuscripts, etc.) should be used when evaluating and further developing the RPN.

\section{Conclusion}
We have introduced Ctrl-F-Net, an end-to-end trainable model for segmentation-free query-by-string word spotting. It simultaneously produces region proposals, and embeds them into a distributed word embedding space in which searches are performed. We outperform the previous state-of-the-art for segmentation-free word spotting on common benchmarks by a large margin and in some cases, even outperforming state-of-the-art segmentation-based approaches. Additionally, we apply Ctrl-F-Net on digitized court records from the early 17\textsuperscript{th} century. This data has not previously been used in word spotting research and serves as a validation that our method is applicable to real-world problems, where pre-segmented words are rare and labeling of training data is expensive.

\section{Acknowledgements}
This project is a part of q2b, From quill to bytes, which is a digital humanities initiative sponsored by the Swedish Research Council (Dnr 2012-5743), Riksbankens Jubileumsfond (NHS14-2068:1) and Uppsala university.


{\small
\bibliographystyle{ieee}
\bibliography{ctrlf_net}
}

\end{document}